# Understanding Unconventional Preprocessors in Deep Convolutional Neural Networks for Face Identification

Chollette C. Olisah[1], Member, IEEE, Lyndon Smith[2],
[1]Baze University, Abuja, Nigeria
[2]The University of the West of England, Bristol, UK

Corresponding author: Chollette C. Olisah (e-mail: chollette.olisah@bazeuniversity.edu.ng).

ABSTRACT Deep convolutional neural networks have achieved huge successes in application domains like object and face recognition. The performance gain is attributed to different facets of the network architecture such as: depth of the convolutional layers, activation function, pooling, batch normalization, forward and back propagation and many more. However, very little emphasis is made on the preprocessors. Unlike other recognition tasks, face recognition calls for careful consideration of the network's input because of image formation factors that can impact on the discriminative characteristics of the face. Therefore, in this paper, the network's preprocessing module is varied across different preprocessing approaches while keeping constant other facets of the deep network architecture, to investigate the contribution preprocessing makes to the network. Commonly used preprocessors are the data augmentation and normalization and are termed conventional preprocessors. Others are termed the unconventional preprocessors, they are: color space converters; HSV, CIE L*a*b* and YCBCR, grey-level resolution preprocessors; full-based and plane-based image quantization, illumination normalization and insensitive feature preprocessing using: histogram equalization (HE), local contrast normalization (LN) and complete face structural pattern (CFSP). To achieve fixed network parameters, CNNs with transfer learning is employed. The aim is to transfer knowledge from the high-level feature vectors of the Inception-V3 network (trained on ImageNet data) to offline preprocessed LFW target data; and features trained using the SoftMax classifier for face identification. The experiments show that the discriminative capability of the deep networks can be improved by preprocessing RGB data with HE, full-based and plane-based quantization, rgbGELog, and YCBCR, preprocessors before feeding it to CNNs. However, for best performance, the right setup of preprocessed data with augmentation and/or normalization is required. Overall, above 72% is achieved with plane-based image quantization which is found to increase the homogeneity of neighborhood pixels and utilizes a more reduced bit depth for better storage efficiency.

INDEX TERMS Deep convolutional neural networks, face identification, preprocessing, transfer learning.

## I. INTRODUCTION

Humans have an intuitive ability to effortlessly analyze, process and store face information for the purposes of identification and authentication [1]. This ability also extends to recognition of face images even at low-resolution [2]. However, since the inception of convolutional neural networks (CNNs), a class of deep machine learning algorithms, machines have been developed that perform face identification and verification tasks at the level of efficiency comparable to humans.

The ability of intelligent machines to perform recognition tasks successfully is dependent on the CNN architecture and the format of the input data. While there is an arsenal of research on the architecture of deep networks [3], there is less focus on the input data of the CNN. This may be due to the perception that the network only needs the raw face images in RGB format in order to extract and learn relevant features for discerning between faces without prior processing. Meanwhile, some recognition applications might present to the network scenarios where information for recognition is not sufficiently represented. For instance, the size of the dataset might be insufficient for training a neural network, distribution of the data may vary, or degradation due to noise or grey-level resolution might be a problem, the color space might differ. Also, there may be cases of intra-person variation resulting from pose differences and/or lighting. Additionally, there may be inter-person similarity where there is close resemblance of faces of persons of different classes. The latter is more typical within a face space than in object space, but it is not within the scope of this study.

Typically, the size of the dataset for training a CNN is highly significant for making sense of the intricate patterns in the data. Compared to traditional machine learning algorithms that employ handcrafted feature extraction algorithms, the amount of data input to the CNN must be large enough, (as is the case for ImageNet [4]. To address the small sample size problem, data augmentation methods such as translation, rotation, scaling and reflection, are often employed in the literature; examples are [5-8]. Another common practice is data normalization for reducing variation in the distribution of data. It involves subtracting the mean from each pixel of the input data and afterwards, the resulting output is divided by the standard deviation of the data. In [9], the study considered the object classification problem. They first adopted the zero-mean and one-standard deviation data normalization method to the training data; then, followed by zero component analysis to enhance the edges of features fed to the CNN. Their work showed significant improvement to the raw-input format. These practices are widely used and are known to substantially improve the CNN performance in face recognition tasks. Another, not so popular, preprocessing strategy, is the color space conversion. Reddy et al. [10] trained the CNN model on different color spaces to investigate influence on the performance of the network. Their results show that the best color format to serve to CNN models as input is the raw RGB image format. Though this study is in the context of object recognition, it appears relevant to face recognition. On the other hand, the robustness of CNN to degraded samples, due to noise, either by study [11] or design [12,13], is gaining interest. The works in [12] and [13] show that image quality affects the convolutional deep networks. Degradation can take another form; image grey-level resolution. This impacts on the visual quality of images due to reduced bit depth. However, reduced bit depth can be useful for real-world applications of face recognition on mobile devices requiring small memory usage and processing. And for this reason, it is a significant problem to be studied.

The problem of dissimilarity due to pose differences and lighting are particularly significant to hand-crafted feature extraction models. With preprocessing approaches such as, face alignment, contrast enhancement, feature preprocessing, etc., employed prior to the facial feature extraction, the face recognition classifiers (after extraction of handcrafted feature processing) showed considerable increases in accuracy [14-16]. Consequently, the effect of some of the commonly used preprocessors were investigated on deep networks. An interesting characteristic of preprocessing in the CNN pipeline can be seen in the work of [17]. Here, raw input images of a general classification problem which includes; places, things and objects, were run through the local contrast normalization pre-processor before sending its output to the network. The pre-processor shows significant improvement to the performance of the model in comparison to the use of raw image format. Pitaloka *et al.* [18] studied different preprocessors, contrast enhancement and additive noising. The results show that preprocessing actually improves recognition accuracy. A remarkable 20.37% and 31.33% CNN performance improvement to the recognition accuracy of the original raw input data were observed with histogram equalization and noise addition, respectively, on facial expression datasets. Hu *et al*. [19] applied and studied various feature preprocessors: large and small-scale feature (LSSF), Difference-of-Gaussian (DOG), and Single Scale Retinex (SSR). These feature preprocessors were applied to face images that were fed to the CNN model. Their work showed that a 10.85% increase to the recognition accuracy of the network can be achieved as compared with the original input data.

In all these studies, the input data, whether as raw or preprocessed data, were trained on the CNN and accuracies of the model were validated and tested. However, the future of deep learning should encompass solving large search engine problems. Search engines, such as googlehouse collections of data on various entities such as, faces, objects, places, things, and so on Therefore it is unclear how deep convolutional models trained on specific recognition task domains, can translate to practical search engine usage given that the data format may have changed. Another challenge is, for a search of a given person's face, for the engine to output like faces of the individual. For cases such as this, transfer learning may be useful. The search engine can house a pool data for features such as, faces, objects, places, and/or things, which vary extensively and are captured at different camera sensitivities. Therefore, it is worth investigating whether preprocessors are relevant when knowledge of general classification feature model parameters is transferred from a pretrained model to a specific new classification problem like face identification.

In this paper, the performance gain of commonly used and yet simple preprocessors is studied from a generalized classification problem to a more streamlined search problem. The preprocessors being considered are chosen with respect to the assumptions made around the data. The assumptions are: a) small sample dataset, b) varying distribution, c) pose variability, d) input data is of a different color space, e) image degradation due to grey-level resolution, f) intra-class dissimilarity due to varying lighting. To overcome these issues, the following category of preprocessors are employed: data augmentation, data normalization, data alignment, RGB to other color space conversion, grey-level quantization, and illumination normalization and insensitive feature preprocessing, respectively. The preprocessed input is then fed to a deep

CNN model pre-trained on 1.2 million images for generalized classification tasks that mimic a search engine application domain. The weights and biases of the pre-trained models are transferred to the new specific search problem, while the last fully-connected layer is updated as new data is trained. The contributions of this paper are summarized as follows:

- Empirical analysis of the preprocessing module in deep networks with knowledge transfer. This is to demonstrate the performance of the network when data format of the target domain is of a different color space, reduced grey-level resolution, or varying lighting, from the source domain.
- Exhaustive evaluation of the conventional preprocessing with unconventional preprocessing methods to investigate the best setup for the preprocessors in deep networks.
- Demonstrate and propose effective preprocessing strategy in deep networks, the plane-based quantization preprocessing. It increases the homogeneity of nearness pixels and also, utilizes a more reduced bit depth for better storage efficiency.

## II. THE FRAMEWORK

The input of the face specific target dataset is passed through preprocessors prior to classifier training. Then, the high-level feature vectors of a generalized data model (comprising of faces, objects, places, things, animals, etc. and trained using a deep CNN model), are transferred to preprocessed data and fed to the classifier. This is to resemble an automated search in a generalized data model search space. This represents a classification problem that is not yet common but is likely to be in the near future. Here, the automated search is focused on face recognition. The setup for the classification task is as illustrated in Fig. 1 and is described in the following subsections.

### A. PREPROCESSING

The raw RGB face images are preprocessed using data augmentation by translation, data alignment by deep funneling, and normalization by zero-mean and one-standard deviation method. Also, HSV, CIELAB and YCBCR color space converters, image quantizers, and illumination normalization and insensitive feature preprocessors using: HE, LN, rgbGELog [20], and CFSP [21]. These preprocessors individually address assumptions a-f, respectively, associated with the dataset.

### 1) DATA AUGMENTATION

Training from scratch or from a pretrained network, requires a good number of samples of data per class for the CNN to generalize well to the given class. Data augmentation has been used as an artificial means to grow the size of the training data. There are different transformation approaches commonly used: translation, rotation, scaling and reflection [5-8]. Each contributes differently to the CNN performance. For interested readers, the work by Paulin *et al.* [22] exemplifies each transformation performance. Since data augmentation is commonly adopted by the deep learning community, it is deemphasized in this work. Therefore, only a translation operation is performed, such that the data are translated by [+30, -30] pixels to create four (4) additional faces shifted to the left, right, top and bottom, per class.

### 2) DATA ALIGNMENT AND NORMALIZATION

A common problem of real-world face image data is that face appearance, from the same person frontal view to his/her profile, varies drastically across large poses. Face alignment is used to improve the performance of handcrafted feature extraction algorithms and is currently applied to input faces to deep networks, to set face data of multiple pose variation to a canonical pose. Since removing the existence of pose variability significantly improves recognition performance, this work continues in the trend by using images aligned by the deep funneling method [23]. Also, to make convergence of the network faster while training, the zero-mean and one-standard deviation is adopted. This ensures that the data input parameters are normalized to exhibit the same data distribution.

### 3) COLOR SPACE CONVERSION

In computer vision, color spaces other than RGB, are somewhat robust to lighting changes. Therefore, following on from the work in [10], this study evaluates the color space preprocessors such as, Hue-Saturation-Value (HSV) and YCBCR. The Y represents luminance, Cb and Cr represent the chrominance component. For the CIE La* b*, L is for luminance, while a* and b* for the green–red and blue–yellow color components. The analysis is not only focused on whether the RGB performance was improved, but also addressed the following question: is it possible for a CNN trained on RGB input data to transfer its knowledge to an input data of different color space?

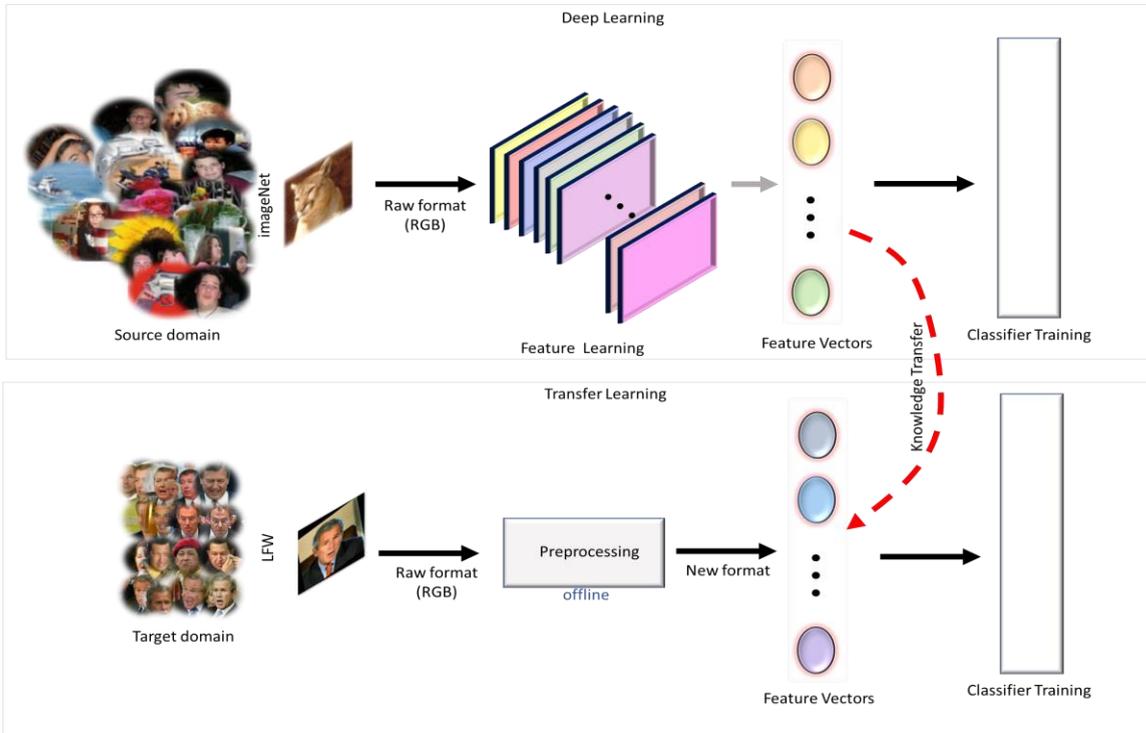

FIGURE 1. The preprocessor module in a deep convolutional learning framework. Fixed network parameters is achieved through transfer of knowledge from generalized raw RGB ImageNet data samples comprising of faces, objects, places, things, animals, etc.to preprocessed target LFW samples.

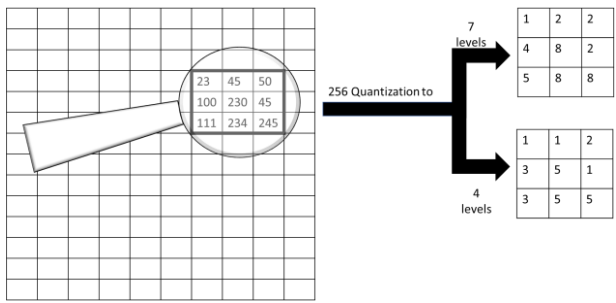

FIGURE 2. Image Quantization. Using a region in an image the homogeneity of nearness pixels is illustrated for a $2^{24}$ grey-level region converted to 7-level and 4-levels, respectively.

### 4) IMAGE DEGRADATION

Grey-level resolution is considered as one of the ways an image can be degraded. The degradation often arises from color quantization which compresses an image grey-level to reduced bit depth. Most online acquired data suffer from color compression because most handheld devices only support a limited number of colors. Two approaches are used: global-based quantization and plane-based quantization. Both employ Otsu's thresholding method, but the former generates a multiple level threshold vector from an RGB image based on a specified level. The threshold vector values change as the quantization level changes. For example, a 6-bits level quantization (see Fig. 2) means that 6-level threshold values are generated from the entire raw format image and used to quantize the image. The latter takes into consideration that the grey-levels changes from plane-to-plane for an RGB image. Therefore, threshold values are generated for the red, green and blue planes to create a 6-threshold vector in 3-planes, (that is, if the choice of quantization level is 6-levels, for quantizing each plane, respectively). To capture real-world mobile devices data, it will be worth observing, the performance of the deep networks with reduced grey-level. However, this study hypothesizes that the quantization process increases the homogeneity of pixels near to each other, which is expected to increase the discriminative capability of the CNN classifier.

### 5) ILUMINATION NORMALIZATION AND INSENSITIVE FEATURE PREPROCESSING

For face images acquired at different spectral bands, the effects of spatial variation in sensitivity of camera systems, is likely to occur. To minimize the variability effect for the face images of the same class, rgbGELog [20] and commonly used illumination normalization techniques are employed. Other approaches, such as LN and CFSP [21], involve the extraction of illumination-insensitive features. These types of feature preprocessors mostly enhance edges as opposed to low-level features. To output a color image with HE, LN and CFSP, a color image version of the preprocessors is used. Given an RGB image, each channel plane is processed individually.

## B. CONVOLUTIONAL NEURAL NETWORKS

Current trends in the application of deep CNN shows that its possibilities in the real-world are endless. A remarkable attribute of deep networks is the ability, with sufficient training data. They can also process raw pixel data directly, extract and learn deep structural features for discrimination and generalize incredibly well to new data. More remarkable is that the deep structural features of the network can be transferred irrespective of the domain [25,26]. For this reason, transfer learning is explored.

## C. TRANSFER LEARNING

A deep CNN architecture designed by Szegedy *et al.* [27], denoted as Inception-V3 was used. The fact that the Inception-V3 model is trained on ImageNet [28], a huge dataset of a million and 200 thousand (1.2 million) generalized data samples and 1000 distinct class labels (of faces, objects, places, things, animals, etc.), makes it a good fit to the objective of this study. It is commonly presented in literature that transfer learning is mainly for situations where training data is insufficient [29]. However, a more significant property of interest, in this study, is the transfer of knowledge from one domain (source) to another, almost unidentical (target), domain. By unidentical it is meant that the data format of the target set might change, either it is of different color space, reduced grey-level resolution, or varying lighting, etc. From the work of [30] the rich features of the CNN at different layers were investigated and their study showed that the lower layers respond to edge-like features, while succeeding layers combine lower layers and more abstract features, which are finally merged at the higher layer as global features. This is likened to the recognition ability of the human visual cortex [1] which processes parts of the face, individually, and are put together as a global feature to make sense of a person's identity. In [31], the output of the last layer, which comprises the high-level feature vectors of a pretrained CNN, showed to generalize well to a new target dataset than fine-tuning some layers of network. It is on this reasons that the high-level feature vectors of the inception-V3 model is found useful to this study's face search problem.

## III. EXPERIMENTAL SETUP

For a better understanding of the experimentation carried out in this study, the data, the transfer model settings and the evaluation strategy are presented in the succeeding subsections.

## A. DATA

The LFW data set [23] is commonly used for modelling real-world data. It is well-known for intra-class variability resulting from pose, illumination and expression problems. The images are in RGB format and comprise of 13233 face images of 5749 individuals. The face search problem with respect to the objective of this study, do not necessarily demand huge data for classifier training. Therefore, only the individuals with over 50 images are considered to enable the classifier to generalize well with the new data.

Each of the deep funneled 1456 [24] face samples belonging to 10-person classes comprises an RGB image of 250 x 250 resolution, but is resized to 299 x 299 resolution because the pretrained network has been trained on images of size 299 x 299 x 3. After resizing, the images are further scaled in the range [0, 1].

## B. TRAINING THE CLASSIFIER

The training was implemented using Inception-V3 pretrained on ImageNet on standard software package, Tensorflow-Slim, as presented in [32]. The lower layers of the network are frozen, while the last (high-level features) layer, believed to resemble human global assemble of individually processed parts of a face, is transferred for training new weights and biases of the face search data. Implementation computation was performed using Intel® Core™ i7-7500U CPU 4 logical processors. The data set was split into training, validation and testing sets, containing 70%, 5% and 25% of the images, respectively. The validation set controls the training process, while the final accuracy is determined using the test set. The Adam optimizer [33] was used with exponential decay, from 0.003 initial learning rate to 0.0001 (exponentially decayed with step down method for every 29 iterations).

## C. EVALUATION

Performance was evaluated using the Top-1 accuracy metric. Since the data augmentation and normalization preprocessors are commonly used on aligned data by the CNN research community, this study terms them conventional preprocessors. Consequently, they become the basis for evaluating the unconventional preprocessors in a CNN as follows: with_augmentation (WA), without_augmentation (NA), with_normalization (WN), and without_normalization (NN). The performance of each of the unconventional preprocessors is reported under these categories: color space conversion, illumination normalization and insensitive feature preprocessors, grey-level resolution degradation.

## IV. RESULTS AND DISCUSSION

Here, the applicability of unconventional preprocessing with conventional preprocessing methods in transfer networks, for a specific target problem such as a face search, is observed and reported. Therefore, the results will be discussed in three stages as follows. Stage one: A general performance of the preprocessors that encompasses: color space conversion, illumination normalization and insensitive feature preprocessors. Stage two: a streamlined report on various grey-level degradation preprocessors. Here various levels of quantization, $2^{24}$ grey-level to 7-levels, 6-levels, 5-levels and 4-levels, are presented. For the third stage, the face search is depicted through a result of the performance of each of the unconventional preprocessors for a search of a given person's identity.

## A. PREPROCESSOR PERFORMANCE IN TRANSFER NETWORK

Table1 shows the evaluation results of passing the target data through preprocessors. The color space conversion, illumination normalization and insensitive feature preprocessors, is reported. The latter is lent from handcrafted feature extraction algorithms. The evaluation of these preprocessors is compared against the raw RGB format commonly used in CNN networks. Therefore, this evaluation seeks to answer the following questions: 1) how much knowledge can be transferred when the input formats of the source and target domains are different? and 2) are preprocessors for addressing the effects of spatial variation due to sensitivity of camera systems relevant in transfer networks?

Surprisingly, the performance of the color space preprocessors showed that format does not hinder knowledge transfer. That is to say that no matter what the input format of a data model is in a transfer setting, the transferred features still generalize well to a new target data. The CIELAB was by far the worst color format across the board. A difference of 28.9772% is observed when compared against the RGB format for the augmented and normalized data experiment. This might be based on the fact that it captures only structural features. Though the RGB has the highest identification accuracy, the YCBCR remained consistent across all the experiments. On average, it outperformed the raw RGB format by a 3.7692% margin.

On the other hand, the illumination normalization and insensitive feature preprocessors appear to have performed similarly, except for the CFSP. The CFSP does not seem to be promising in a transfer network. The best performing illumination preprocessor is the HE followed by LN.

Additionally, it is obvious that different preprocessors react differently with data augmentation and normalization. The RGB, HE, CFSP, and LN perform better when data are normalized and augmented. The YCBCR is best with only data augmentation and rgbGELog is best with only normalized data.

## B. GREY-LEVEL PREPROCESSOR PERFORMANCE IN TRANSFER NETWORK

Even more interesting is the performance of grey-level resolution reduction preprocessors in transfer networks. As shown in Table 2, the results of this experiment verifies our claim that homogeneity of nearness pixels is increased through quantization. A 72.7273 Top-1 accuracy is achieved for a $2^{24}$ grey-level image quantized to 7-levels using the plane-based quantization approach. Also, the experiment shows that creating variants of given data by translation does not favor quantized data in deep networks. Despite the fact that the raw RGB format data achieved a 70.7386% accuracy with increased data, it is worth noting that the best accuracy is in fact achieved without augmentation.

In the first experiment, with data augmentation and normalization, the quantization to 7-levels using the full-image based approach outperforms its other counterparts as well as the plane-based quantization approach. The same performance is achieved for data quantized to 4-levels with no augmentation and normalization.

Also, in the second and third experiments, the 5-level quantization preprocessors competed between augmentation and normalization with almost a draw in performance. However, the 7-level quantization performs significantly well when data is normalized.

## C. THE TRANSFER NETWORK FACE SEARCH PROBLEM

The result of this experiment is reported in Fig. 3 and Table 3. Given the different types of preprocessors, of which their individual performances have been observed, it is important to determine the practicability of the face search in the transfer network. The transfer network makes the search problem solvable. Assuming a large database of faces, objects, places, things, and so on, a search engine searching for a face identity might take hours and days. This is when the transfer network becomes really useful, because the engine can be modelled to comprise a large engine (for all the possible classes that comprises of face, objects, places, things, etc.) and a mini-engine (specific sample data), possibly of a handheld mobile device. Here it is of interest to know how the preprocessors are able to retrieve the identity of a given face relative to other face classes of the target dataset. This experiment is presented in Fig. 4.

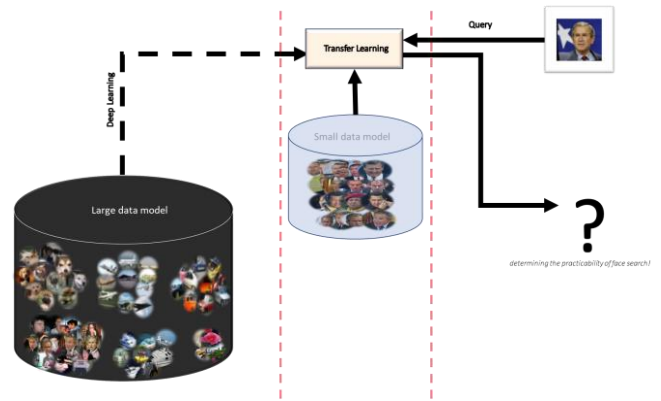

Figure 3. The face search data model. Exploring the identity retrieval of different preprocessors in a transfer network.

Fig. 4 shows the result of querying the engine for Ariel Sharon and George W. Bush to determine their identity. It appears that the CNN network with plane-based quantization from $2^{24}$ grey-level to 7-levels achieved a 99.88% accuracy with no augmentation and normalization while the CNN network with RGB format achieved 99.27% with augmentation and normalization.

TABLE II
COMPARING THE PERFORMANCE OF RGB DATA AND GREY-LEVEL PREPROCESSORS IN A TRANSFER NETWORK

| Category | Preprocessor | WA & WN (%) | WA & NN (%) | NA & WN (%) | NA & NN (%) | Mean (%) |
|---|---|---|---|---|---|---|
| Original | RGB | 70.7386 | 65.6250 | 51.1364 | 50.0000 | 59.3700 |
| | F-4-Level | 66.4773 | 63.9205 | 68.4659 | 66.4773 | 66.3353 |
| | F-5-Level | 63.3523 | 64.4886 | 65.0568 | 64.7727 | 64.4176 |
| Full | F-6-Level | 63.9205 | 60.2273 | 62.2159 | 59.0909 | 61.3637 |
| | F-7-Level | 67.6136 | 69.6023 | 67.0455 | 67.0455 | 67.8262 |
| | P-4-Level | 64.4886 | 63.3523 | 63.3523 | 67.6136 | 67.7017 |
| Plane | P-5-Level | 64.7727 | 70.1705 | 70.4545 | 65.0568 | 67.6136 |
| | P-6-Level | 65.6250 | 64.2045 | 67.6136 | 63.3523 | 65.1989 |
| | P-7-Level | 67.0455 | 68.7500 | 72.7273 | 65.3409 | 68.4659 |

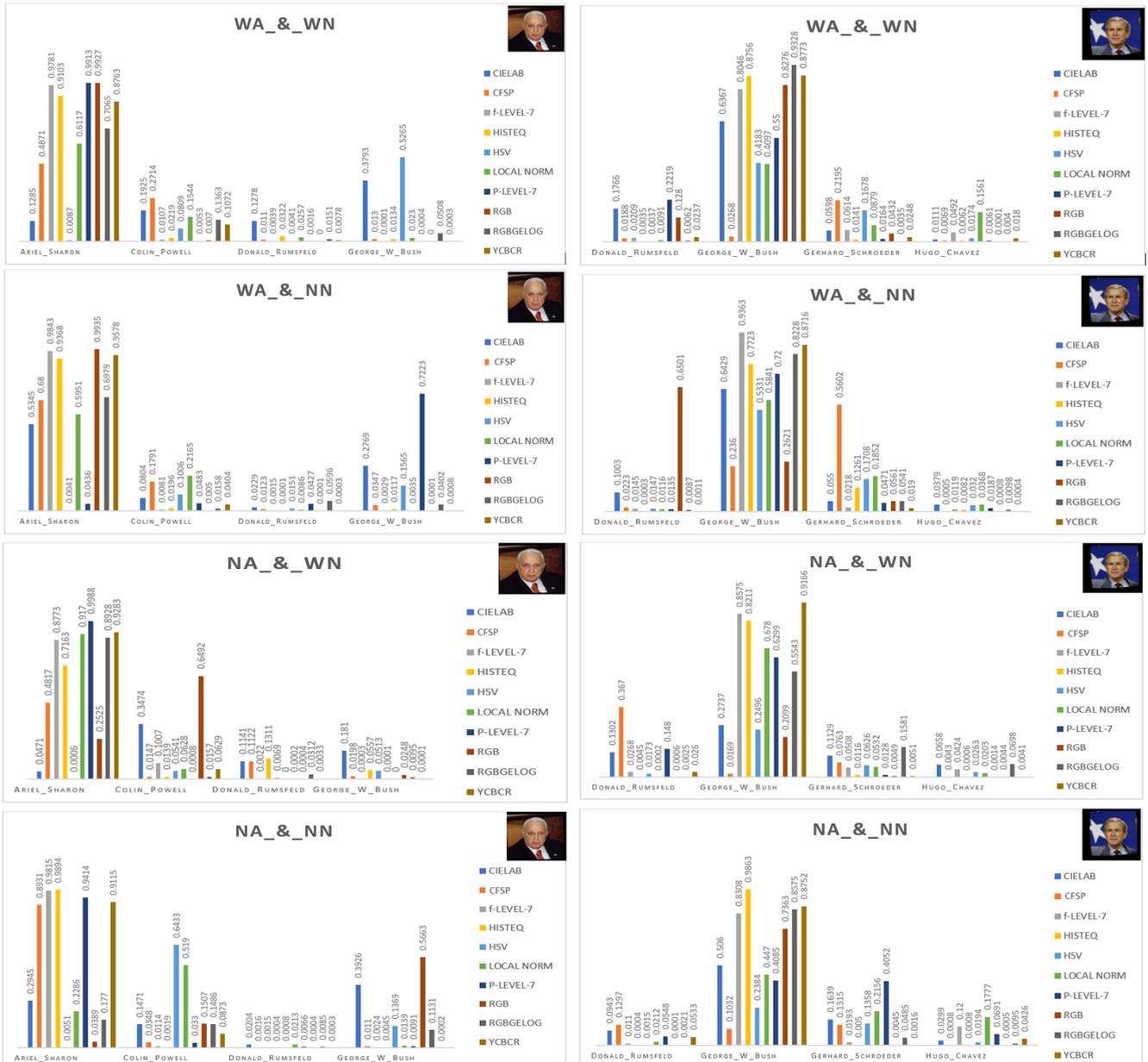

Figure 4. Identity retrieval accuracy for Ariel Sharon and George W. Bush across the 10 face classes for different preprocessors. The view is limited to top 4 accuracies of the identity retrieval.

Surprisingly, the HE, YCBCR, rgbGELog, full and plane-based quantization preprocessors were equally useful and competitive at retrieving the right identity of a query image. However, in order to attain the best performance of any of these preprocessors, it is best to utilize them based on their acceptance for some conventional preprocessing such as augmentation and normalization.

The face search experiment also reveals that the raw RGB format only performs well when data size is increased by augmentation and the distribution of data is normalized. Other than this, it fails to perform favorably. The YCBCR, rgbGELog, and HE maintained good accuracies independent of normalization and or augmentation.

V. CONCLUSION

In contrast to what is commonly believed, the preprocessor module for deep learning frameworks has proved significant in deep networks. In this paper, the various facets of the CNN architecture were kept constant while the preprocessing module was varied for different preprocessing algorithms. The HE, full-based and plane-based quantization, rgbGELog, and YCBCR (these are the unconventional preprocessors in CNNs), showed that the discriminative capability of the deep networks can be improved by preprocessing the raw RGB data prior to feeding it to the network. However, the best performance of these preprocessors was achieved via considering, under various preprocessing setups, data augmentation and/or normalization (these are the conventional preprocessors) in CNNs. Even though the raw RGB format performed well, quantizing a $2^{24}$ grey-level image to 7-levels outperformed the RGB and achieved above 72% accuracy with data normalization. This preprocessor was found to be an effective preprocessing strategy in deep networks for the reasons that it might have increased the homogeneity of neighborhood pixels and utilizes a more reduced bit depth for better storage efficiency.

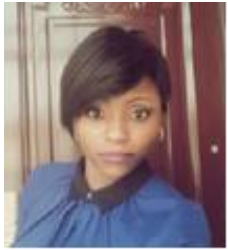

CHOLLETTE C. OLISAH received the B.Sc. degree in Computer Science from Anambra State University, Uli. She received the M.Sc. and PhD degrees in Computer Science from the Universiti Teknologi Malaysia in 2011 and 2015, respectively. From 2016 till present she has been a lecturer with the Department of Computer Science, Baze University, Abuja, Nigeria. She has authored over 12 peer-reviewed scholarly, conference and journal, articles. Her research interests are in machine learning (particularly, neural networks and deep learning), image processing, face recognition, image understanding and analysis. Dr. Chollette has served as reviewer for reputable journals, from 2015 to till date.

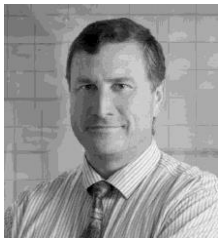

LYNDON SMITH is Professor in Computer Simulation and Machine Vision at the University of the West of England, Bristol (UWE), UK. He has twenty-five years of experience of research (on both sides of the Atlantic), with an emphasis on 3D vision for analysis of complex surface textures and object morphologies; and a particular interest in modelling using neural networks and deep learning. His research interests included development of the 3D Skin Analyser for analysis of potentially cancerous skin lesions; and a new method for 3D computer simulations of irregular morphologies. He has successfully supervised twenty PhDs and is currently engaged in the directorship of the UWE Centre for Machine Vision. Professor Smith has also published 180 papers, written a book, edited another book, undertaken 7 guest editorships of international journals, has been named as inventor in 12 patent applications and has developed working prototype and commercial machine vision systems that have been delivered to clients internationally.